\documentclass{article}

\usepackage{microtype}
\usepackage{graphicx}
\usepackage{subcaption}
\usepackage{booktabs} 
\usepackage{amsmath}
\usepackage{amssymb}
\usepackage{natbib}

\usepackage{hyperref}

\usepackage[accepted]{icml2019}

\icmltitlerunning{MISO: Mutual Information Loss with Stochastic Style Representations for Multimodal Image-to-Image Translation}

\usepackage{cuted}
\usepackage{capt-of}

\begin{document}

\twocolumn[
\icmltitle{MISO: Mutual Information Loss with Stochastic Style Representations \\for Multimodal Image-to-Image Translation}





\begin{icmlauthorlist}
\icmlauthor{Sanghyeon Na}{KU}
\icmlauthor{Seungjoo Yoo}{KU}
\icmlauthor{Jaegul Choo}{KU}
\end{icmlauthorlist}

\icmlaffiliation{KU}{Department of Computation, Korea University, Seoul, Korea}

\icmlcorrespondingauthor{Sanghyeon Na}{sktkdgus98@korea.ac.kr}
\icmlcorrespondingauthor{Seungjoo Yoo}{seungjooyoo@korea.ac.kr}
\icmlcorrespondingauthor{Jaegul Choo}{jgchoo@korea.ac.kr}



\vskip 0.3in
]




\begin{abstract}
Unpaired multimodal image-to-image translation is a task of translating a given image in a source domain into diverse images in the target domain, overcoming the limitation of one-to-one mapping. Existing multimodal translation models are mainly based on the disentangled representations with an image reconstruction loss. We propose two approaches to improve multimodal translation quality. First, we use a content representation from the source domain conditioned on a style representation from the target domain. Second, rather than using a typical image reconstruction loss, we design MILO (Mutual Information LOss), a new stochastically-defined loss function based on information theory. This loss function directly reflects the interpretation of latent variables as a random variable. We show that our proposed model Mutual Information with StOchastic Style Representation(MISO) achieves state-of-the-art performance through extensive experiments on various real-world datasets.
\end{abstract}  

\begin{figure}\centering
\includegraphics[width=\columnwidth]{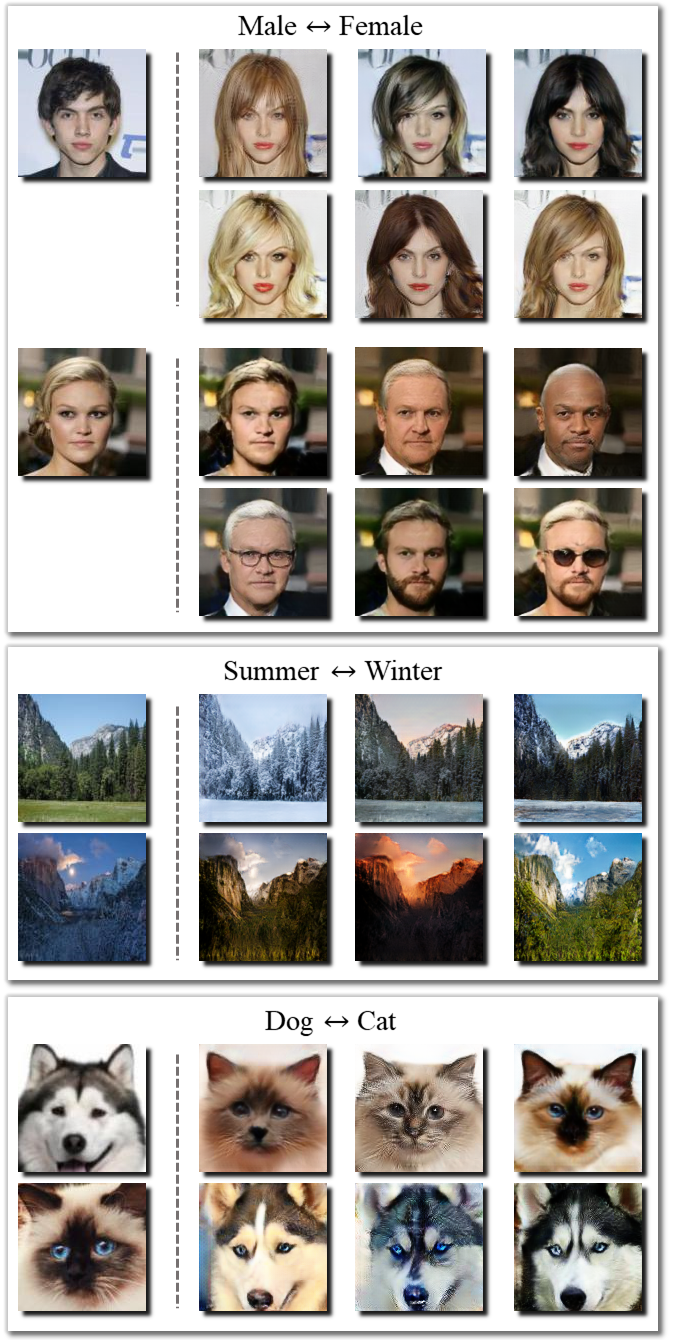}
\captionof{figure}{\textbf{Multimodal image-to-image translation on multiple datasets}. MISO can produce high-quality and diverse images for a single input image.} 
\label{fig:teaser}
\end{figure} 

\section{Introduction} \label{sec:intro}
Unpaired image-to-image translation is a task of translating an image belonging to one domain into an image in another domain without aligned data across the two domains. Multimodality means that a single image can be translated into many different images in the other domain. It is a fundamental issue that makes this task challenging. Real-world data is inherently multimodal, as the man in the first row of Fig.~\ref{fig:teaser} can be imagined as many different women. Unpaired multimodal image-to-image translation(multimodal translation) aims to incorporate this property and generate diverse images from a single input image. 

However, multimodal translation is a complicated task that should be handled differently from one-to-one translation. We propose two assumptions that can better handle the properties of multimodality. 
First, we propose that there exists a hierarchy between content and style. Instead of strictly dividing content and style as independent features during the disentanglement process like existing models, we assume that content should be the base and style should be conditioned to the content to perform a high-quality translation. 
Second, self-reconstruction(SR) loss which is utilized by existing multimodal translation models
is not the best for multimodal translation. Therefore, we present a new probabilistic loss function that can replace the SR loss.

To accomplish multimodal translation, we need to learn one-to-many mapping rather than one-to-one mapping between the two domains. It can be achieved by learning a mapping from a pair of source image and random noise to a target image. To learn this mapping effectively, BicycleGAN~\cite{zhu2017toward} proposes two-phase training which consists of $\mathcal{X} \rightarrow \mathcal{Z} \rightarrow \mathcal{X}$ (Image-Feature-Image, IFI) and $\mathcal{Z} \rightarrow \mathcal{X} \rightarrow  \mathcal{Z}$ (Feature-Image-Feature, FIF) where $\mathcal{X}$ is an image space and $\mathcal{Z}$ is a feature space. Each of these phases has core loss function. However, it is a paired translation model.

MUNIT~\cite{huang2018multimodal} and DRIT~\cite{lee2018diverse} are models that previously attempted to solve unpaired multimodal translation. They adopt two-phase training by explicitly disentangling content features and style features. An example of content and style can be found in face datasets that have two domains; male and female. The content features, also called domain-invariant (DI) features, could be the background, angle of face and gaze. Style features, which are also called domain-specific (DS) features, include features that are unique to each domain such as long hair and makeup for the female domain and beards for the male domain. As IFI loss, both models use a SR loss which is a L1 loss between the source image and its reconstructed image in the encoder-decoder structure.

Although our model also follows the two-phase training of previous models, we use the content feature conditioned on the style feature rather than the independent content and style feature. This is because we assume that content feature from the source image should be maintained and the style feature should slightly cover the content feature for a successful style transfer. In this end, the conditional encoder learns how to condition the content feature with the style feature to generate high-quality images. 
In addition, we propose a new loss function for multimodal translation replacing the SR loss. The SR loss does not capture the detailed characteristics of the image, resulting in a lack of delicate diversity in multimodal translation. To overcome this limitation, we interpret the latent variables as a random variable and define a new stochastic loss function based on mutual information to reflect this interpretation. With a different perspective on a relationship between content and style feature and a new loss function replacing the SR loss, we build a outperforming model in terms of reality and diversity. We demonstrate the effectiveness of our models extensively using quantitative and qualitative experiments.

\begin{table}[t]
\centering
\caption{\textbf{Comparisons of MISO with previous models}. IFI-loss:self-reconstruction(SR) or mutual information(MI). F-hierarchy: whether there is a hierarchy between content and style features. Latent var: Interpreting latent variable as deterministic value (DV) or random variable (RV).}
\vskip 0.1in
\label{baseline_comparison}
\begin{center}
\begin{small}
\begin{sc}
\begin{tabular}{c|ccc}
\toprule
 & MUNIT & DRIT & MISO(Ours) \\
\midrule
IFI-loss  & sr & sr & mi \\
F-Hierarchy & $\times$ & $\times$ & $\surd$ \\
latent var  & DV & RV & RV \\
\bottomrule
\end{tabular}
\end{sc}
\end{small}
\end{center}
\vskip -0.1in
\end{table}

\section{Related Work}
\begin{figure*}
        \begin{subfigure}[b]{0.6\textwidth} 
                \includegraphics[width=\linewidth]{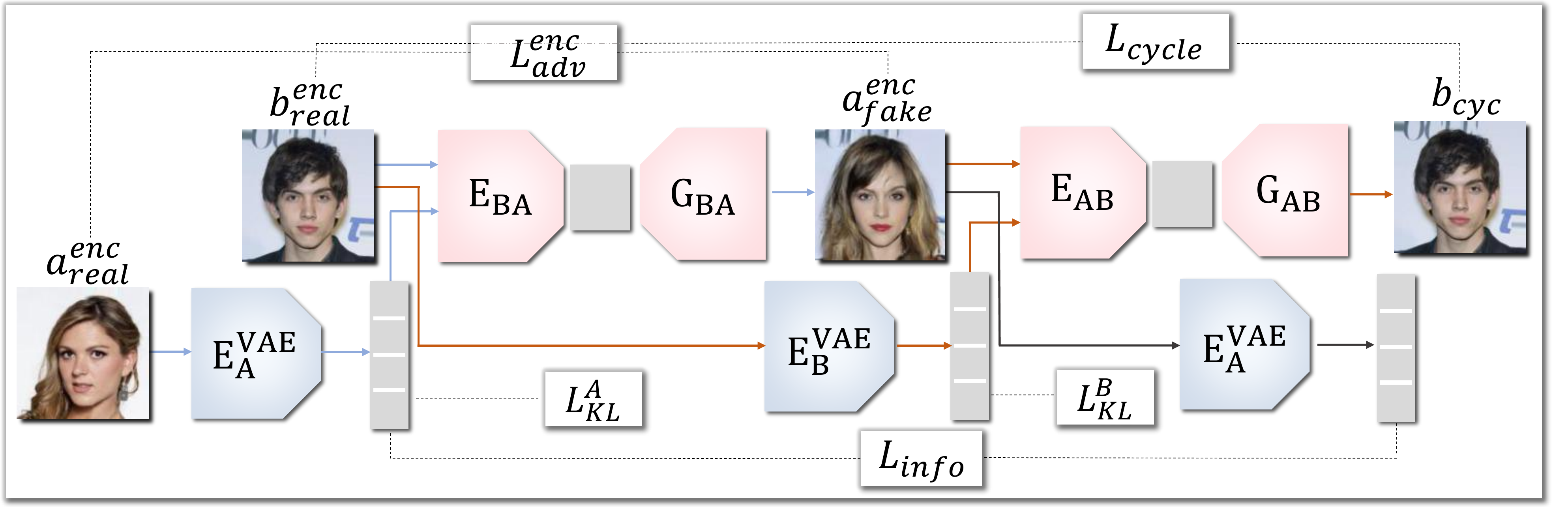}
                \caption{IFI phase}
                \label{subfig:phase_1}
        \end{subfigure}%
        \begin{subfigure}[b]{0.4\textwidth}
                \includegraphics[width=\linewidth]{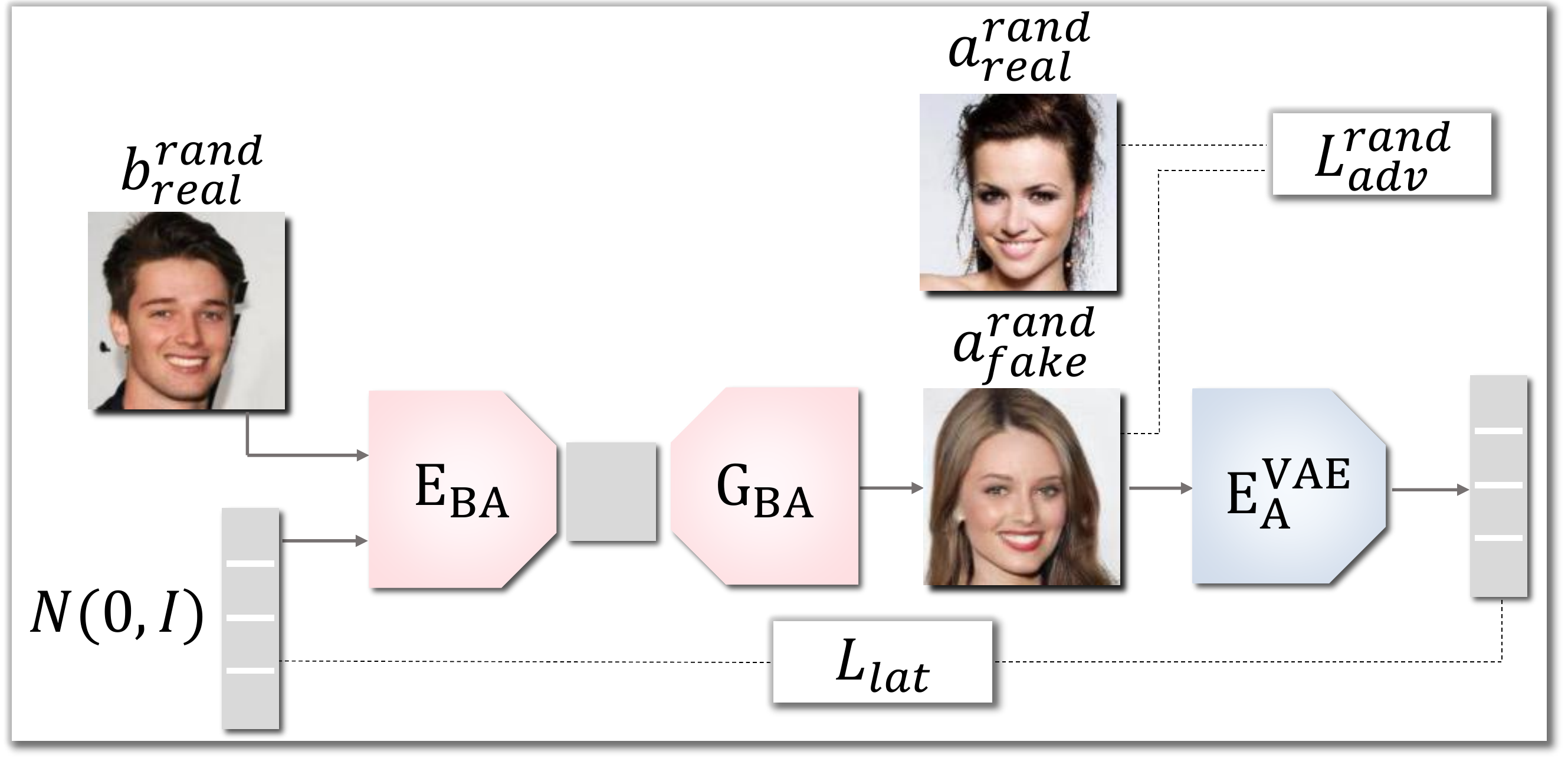}
                \caption{FIF phase}
                \label{subfig:phase_2}
        \end{subfigure}%
        \caption{Model architecture of our proposed model. Eq. ~\eqref{eq:info_loss}, Eq.~\eqref{eq:adv_enc_loss}, Eq.~ \eqref{eq:cycle_loss}, Eq.~\eqref{eq:KL_A_loss} and Eq.~\eqref{eq:KL_B_loss} are used in (a) and Eq. \eqref{eq:adv_rand_loss} and Eq.~\eqref{eq:lat_recon_loss} in (b). Note that we use different images for each phase.}\label{fig:model_fig}
\end{figure*}

\subsection{Mutual Information}
In information theory, mutual information measures how dependent two random variables are. Mutual information $I$ between two variables $X$ and $Y$ is defined as
\begin{equation}
I(X;Y) = \int_y \int_x p(x,y) \log\left(\frac{p(x,y)}{p(x)p(y)}\right) dx dy. \nonumber
\end{equation}
MINE~\cite{belghazi2018mine} proposes neural estimator of mutual information to improve generative models.
Measuring mutual information usually requires the true posterior $p(y|x)$ which makes it intractable. InfoGAN~\cite{chen2016infogan} introduces a lower bound on mutual information through Variational Information Maximization~\cite{agakov2004algorithm}. To the best of our knowledge, our model is the first model to use mutual information to introduce a stochastic loss function for multimodal translation.

\subsection{Multimodal Image-to-Image Translation} \label{subsec:mul_img2img}
Image-to-image translation has made significant progress in super-resolution~\cite{ledig2017photo}, colorization~\cite{bahng2018coloring} or inpainting~\cite{yeh2017semantic} with the advent of generative adversarial networks (GANs)~\cite{goodfellow2014generative} and other methods to stabilize the training of GANs~\cite{radford2015unsupervised, arjovsky2017wasserstein, gulrajani2017improved, miyato2018spectral, metz2016unrolled, tran2018dist} and generate high-resolution images~\cite{karras2017progressive, zhang2018self, brock2018large, karras2018style}.
Early models~\cite{yi2017dualgan,CycleGAN2017,kim2017learning,liu2017unsupervised} can only generate a single output for a single input. Multi-domain image-to-image translation~\cite{StarGAN2018,anoosheh2017combogan} proposes new methods to generate diverse outputs with additional domain labels. BicycleGAN~\cite{zhu2017toward} first proposes two-phase  training using paired data to produce multimodal outputs without additional information. MUNIT~\cite{lee2018diverse} and DRIT~\cite{huang2018multimodal} extends this task into the unpaired setting. Note that Augmented CycleGAN~\cite{almahairi2018augmented} is excluded from the baselines as reproducing reasonable results is not possible. These models work under a disentanglement assumption, which assumes that content feature space and style feature space can be separated and different domains share the same content space. Since this assumption does not put enough importance on content features, generated outputs can lose identity of the content image. That is why our model introduces a hierarchy between the two features for a high-quality output. Also both models use SR loss. Previous works point out that this SR loss fails to capture detail features~\cite{larsen2015autoencoding, isola2017image} because pixels may be blindly averaged out, resulting in a blurry output. Therefore, we propose a new loss function to learn better features for the multimodal translation problem which can effectively replace the SR loss.

\section{Method} \label{sec:method}

We introduce our unpaired multimodal image-to-image translation model MISO and its components. Our goal is to learn a one-to-many mapping between two domains $A \subset \mathbb{R}^{H \times W \times 3}$ and $B \subset \mathbb{R}^{H \times W \times 3}$. These two domains can be used interchangeably as source domain $\mathcal{S}$ and target domain $\mathcal{T}$. One-to-many mapping between $\mathcal{S}$ and $\mathcal{T}$ can be achieved by learning distribution $p(t|s,z)$ where $t \in \mathcal{T}$, $s \in \mathcal{S}$ and $z \sim \mathcal{N}(0, I)$. In other words, our model needs to learn a one-to-one mapping of $(\mathcal{S}, \mathcal{Z}) \mapsto \mathcal{T}$ where $z \in \mathcal{Z}$. It is important to note that $z \sim \mathcal{N}(0, I)$ does not have any power to force a particular $s$ to be mapped to a particular $t$. Therefore, we use an encoder to extract a feature $z_t$ from $t$ and make an arbitrary $z$ obtain the information in $z_t$. 
Finally, we can model a distribution of $p(t|s,z)$ where $z \sim \mathcal{N}(0, I)$. After the training process, we can generate diverse images from a single image by sampling from $p(t|s,z)$ where $z \sim \mathcal{N}(0, I)$.

Fig.\thinspace\ref{fig:model_fig} shows the training process of MISO. Style is extracted from image $a$ of domain A and content is extracted from image $b$ of domain $B$. Though not shown in the figure, the same process is carried out with swapped domains. 

Our model consists of two style encoders for each domain($E_A:A \mapsto Z_A$ and $E_B:B \mapsto Z_B$), two discriminators for each domain($D_A:A \mapsto \mathbb{R}$ and $D_B:B \mapsto \mathbb{R}$), two conditional encoders for each direction($E_{AB}:(A, Z_B) \mapsto Z_{AB}$ 
and $E_{BA}:(B, Z_A) \mapsto Z_{BA}$) and two generators for each direction($G_{AB}:Z_{AB} \mapsto B$ and $G_{BA}:Z_{BA} \mapsto A$). For brevity, we can represent $E_{AB}$ and $G_{AB}$ together as $G_{AB}$ and $E_{BA}$ and $G_{BA}$ as $G_{BA}$.

Our encoders are based on variational autoencoder(VAE)~\cite{kingma2013auto} architecture because we do not assume deterministic mappings for $E_A:A \mapsto Z_A$ and $E_B:B \mapsto Z_B$. VAE architecture can handle the intractable true posteriors $p(z_a|a)$ and $p(z_b|b)$ via their approximated posteriors $q(z_a|a)$ and $q(z_b|b)$, which can be regarded as normal distributions, where $a \in A$, $b \in B$, $z_A \in Z_A$ and $z_B \in Z_B$.

\subsection{MILO: Mutual Information Loss} \label{subsec:info_loss}
This section introduces our proposed loss function Mutual Information LOss(MILO) to replace SR loss. Our motivation starts from considering the goal of multimodal translation as learning the distribution of $p(t|s,z)$. It would be better to regard $z$ as a random variable which has a true posterior of $p(z|x)$ where $x \in \mathcal{X}$ rather than a deterministic mapping of $\mathcal{X} \mapsto \mathcal{Z}$ because our goal is modeling the distribution $p(t|s,z)$. This perspective gives a randomness to the feature extracted from a single image which means that we can assign a different weight to the characteristics that the image has whenever the feature is extracted. Furthermore, this perspective gives a distribution of encoded latent variables which does not exist in the deterministic values. By minimizing the distance between the encoded latent distribution and $\mathcal{N}(0, I)$, a random vector sampled from $\mathcal{N}(0, I)$ can obtain information about the target domain. 

To reflect this randomness, we design our stochastically-defined loss function MILO, that maximizes mutual information between feature $E_A(a)$ and image generated with that feature $G_{BA}(b, E_A(a))$, written as $I(E_A(a);G_{BA}(b, E_A(a)))$. To this end, the generator should use $E_A(a)$ in generating $G_{BA}(b, E_A(a))$ to maximize mutual information. For brevity, $I(E_A(a);G_{BA}(b, E_A(a)))$ can be written as $I(z_a;G(b,z_a))$. Mutual information is hard to directly be maximized, so we induce a lower bound on the mutual information inspired by InfoGAN using an approximation of the true posterior as
\begin{figure*}
        \begin{subfigure}[b]{0.5\textwidth}
                \includegraphics[width=\linewidth]{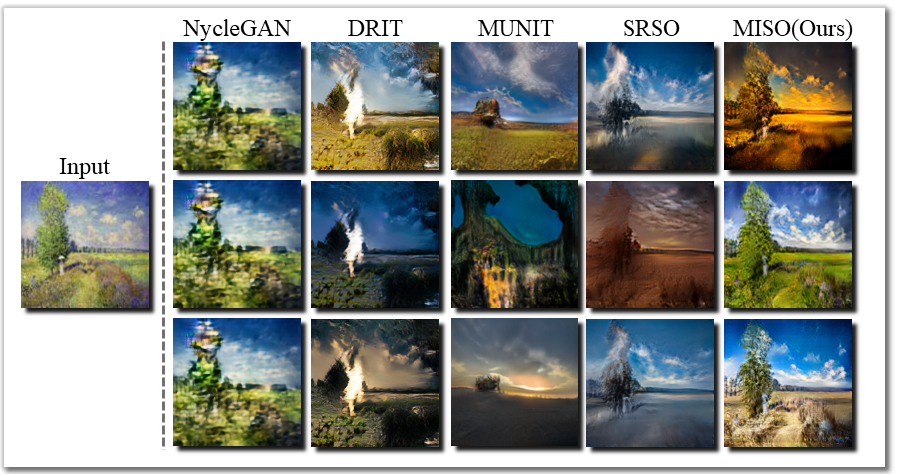}
                \caption{Monet $\rightarrow$ Photo}
                \label{subfig:comparison_sub_1}
        \end{subfigure}%
        \begin{subfigure}[b]{0.5\textwidth}
                \includegraphics[width=\linewidth]{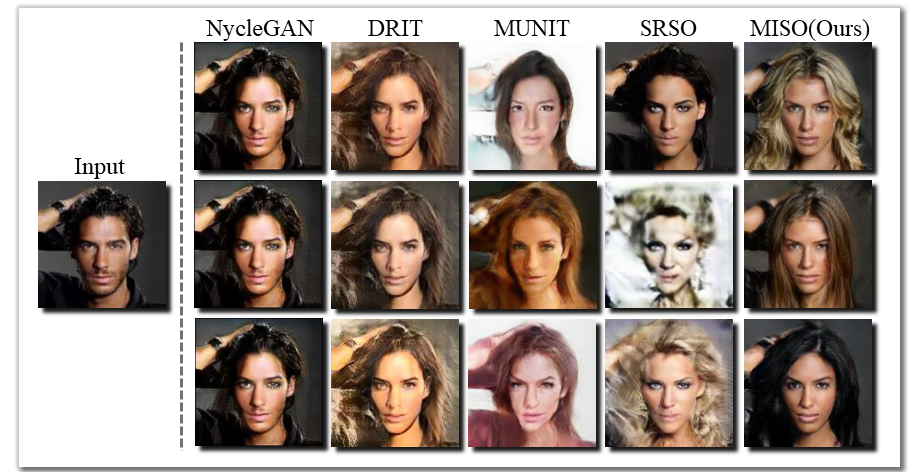}
                \caption{Male $\rightarrow$ Female}
                \label{subfig:comparison_sub_2}
        \end{subfigure}%
        \caption{\textbf{Comparison to baselines.} (a) We can see that our model succeeds at maintaining content of the source image. (b) Existing models fail to show meaningful diversity when given human face images that require intricate translation. Our model (the rightmost column) is able to produce diverse, high-quality outputs even on human faces.}\label{fig:comparison_sampling}
\end{figure*} 
\begin{align}
&I\left( z_a;\thinspace G(b,z_a)\right) \nonumber \\
& = H(z_a)-H(z_a|G(b,z_a)) \nonumber \\
& \ge \mathbb{E} \thinspace _{z_a \sim \thinspace p(z_a)} [\mathbb{E} \thinspace _{a' \sim \thinspace p_g(a|(b, z_a))}[\log q(z_a|a')]] + H(z_a) \label{eq:first_eq}
\end{align}  
where $H$ is an entropy, $q(z_a|a')$ is an approximated distribution(or the encoded latent distribution) of the intractable true posterior $p(z_a|a')$ and $p_g$ is a distribution of the generated image $G(b, z_a)$ called generative distribution. Furthermore, the conditional encoder is a deterministic function, so we include mapping of $(b, z_a) \mapsto z_{ba}$ in the sampling process from the generative distribution $p_g$.
In Eq.\thinspace\eqref{eq:first_eq}, $b$ is a source image sampled from $p(b)$ so we can treat it as a constant. In addition, we can treat $H(z_a)$ as a constant by fixing the distribution of $z_a$. Thus, we can rewrite Eq.\thinspace\eqref{eq:first_eq} as
\begin{equation}  \label{eq:second_eq}
    \mathbb{E} \thinspace _{z_a \sim \thinspace p(z_a)}[\mathbb{E} \thinspace _{a' \sim p_g(a|z_a)}[\log q(z_a|a')]].
\end{equation}

However, we have an unobserved prior of a latent variable, $p(z_a)$ that we cannot directly sample from. We want to make sure that although we generate images with $z \sim \mathcal{N}(0, I)$ this does not mean $\mathcal{N}(0, I)$ is a prior of $z_a$. $\mathcal{N}(0, I)$ should be regarded as an arbitrary distribution close to the encoded latent distribution. The distance between encoded latent distribution and $\mathcal{N}(0, I)$ is minimized to use $z \sim \mathcal{N}(0, I)$ at the inference time by the KL-divergence loss (Section~\ref{subsec:KL_loss}).
We can bypass the problem of sampling directly from $p(z_a)$ by using another distribution $p(a)$ that we can easily sample from. This leads to rewriting Eq.~\eqref{eq:second_eq} as
\begin{align} \label{eq:third_eq}
&\int_{z_a} \mathbb{E} \thinspace _{a' \sim p_g(a|z_a)}[\log q(z_a|a')] p(z_a)\, dz_a \nonumber \\
& = \int_{z_a} \int_{a} \mathbb{E} \thinspace _{a' \sim p_g(a|z_a)}[\log q(z_a|a')] p(z_a, a)\, da\, dz_a \nonumber \\
& = \int_{a} \int_{z_a} \mathbb{E} \thinspace _{a' \sim p_g(a|z_a)}[\log q(z_a|a')] p(z_a, a)\, dz_a\, da \nonumber \\
& = \mathbb{E} \thinspace _{a \sim p(a)}[\mathbb{E} \thinspace _{z_a \sim p(z_a|a)}[\mathbb{E} \thinspace _{a' \sim p_g(a|z_a)}[\log q(z_a|a')]]].
\end{align}
A problem arises with the term $z_a \sim \thinspace p(z_a|a)$ in Eq.~\eqref{eq:third_eq} as this means sampling from the true posterior which is impossible. However, we can use $q(z_a|a)$ which is an approximation of $p(z_a|a)$. We now obtain the final form of the lower bound of $I(z_a; \thinspace G(b,z_a))$ as
\begin{align} \label{eq:fourth_eq}
I(z_a; & \thinspace G(b,z_a)) \nonumber \\
& \ge \mathbb{E} \thinspace _{a \sim p(a)}[\mathbb{E} \thinspace _{z_a \sim q(z_a|a)}[\mathbb{E} \thinspace _{a' \sim p_g(a|z_a)}[\log q(z_a|a')]]].
\end{align}
One can consider $q(z_a|a')$ in Eq.~\eqref{eq:fourth_eq} as a normal distribution as we use a VAE-style encoder, and thus $q(z_a|a')$ can be represented as $\mathcal{N}(\thinspace \mu_{out}, \sigma_{out}^2)$ where $\mu_{out}$ and $\sigma_{out}$ are outputs of the encoder given $a'$ as an input. 
As $\mathcal{N}(\thinspace \mu_{out}, \sigma_{out}^2)$ has a closed form of the probability density function $f(z_a | \mu_{out}, \sigma_{out}^2)$, we can represent $\log f(z_a| \mu_{out}, \sigma_{out}^2)$ as 
\begin{align}
\log f(z_a| \mu_{out}, \sigma_{out}^2) 
& = -\frac{1}{2}\log 2 \pi \sigma_{out}^2 - \frac{(z_a - \mu_{out})^2}{2 \sigma_{out}^2}. \nonumber
\end{align}
To maximize $I(z_a; \thinspace G(b,z_a))$, $\log f(z_a| \mu_{out}, \sigma_{out}^2)$ should also be maximized. Finally, MILO(denoted $\mathcal{L}_{info}$) to be minimized is defined as
\begin{equation} \label{eq:info_loss}
    \mathcal{L}_{info} = \frac{1}{2}\log 2 \pi \sigma_{out}^2 + \frac{(z_a - \mu_{out})^2}{2 \sigma_{out}^2}.
\end{equation}

\subsection{Adversarial Loss} \label{subsec:adv_loss}
To make the generated images indistinguishable from the real images, we employ the adversarial loss in the GAN framework. Note that one can use both of $z \sim \mathcal{N}(0, I)$ and $z_a \sim q(z_a|a)$ when translating images. Although these two distributions will be close to each other after training, they are not identical. Both $a_{fake}^{enc}$ and $a_{fake}^{rand}$ are needed to guarantee high-quality outputs. The adversarial losses for both of them are defined as

\begin{equation} \label{eq:adv_enc_loss}
\begin{split}
\mathcal{L}_{adv}^{enc} 
= & \thinspace \mathbb{E} \thinspace _{a \sim p(a)}[\log D_A(a)] \\
& + \mathbb{E} \thinspace _{b \sim p(b), \, z_a \sim q(z_a|a)}[\log (1-D_A(G_{BA}(b, z_a)))],
\end{split}
\end{equation}
\begin{equation} \label{eq:adv_rand_loss}
\begin{split}
\mathcal{L}_{adv}^{rand} 
= & \thinspace \mathbb{E} \thinspace _{a \sim p(a)}[\log D_A(a)] \\
& + \mathbb{E} \thinspace _{b \sim p(b), \, z \sim \mathcal{N}(0, I)}[\log (1-D_A(G_{BA}(b, z)))].
\end{split}
\end{equation}
Finally, the full adversarial loss can be written as
\begin{equation}
\mathcal{L}_{adv} = \mathcal{L}_{adv}^{enc} + \mathcal{L}_{adv}^{rand}. 
\end{equation}
Equal weights are assigned to both losses.

\subsection{Cycle-Consistency Loss} \label{subsec:cycle_loss}
It is important to preserve content of the source image and only change its style. Significantly altering content of an input image will result in generating a completely different image which defeats the purpose of style transfer. Thus, cycle-consistency loss~\cite{CycleGAN2017} is included for this task, i.e., 
\begin{equation} \label{eq:cycle_loss}
\mathcal{L}_{cyc} = \mathbb{E} \thinspace _{a \sim p(a), \thinspace b \sim p(b)}[|| \thinspace G_{AB}(\bar{a}, E_B(b)) \thinspace - \thinspace b \thinspace ||_1]
\end{equation}
where $\bar{a} := G_{BA}(b, E_A(a))$ corresponds to fake image of domain A(denoted $a_{fake}^{enc}$) in Fig.\thinspace\ref{subfig:phase_1}. To reconstruct $b$, $\bar{a}$ should contain content feature of $b$. Existing models such as DRIT obtain style features($E_B(b)$ in Eq. \eqref{eq:cycle_loss}) from generated images which can be impaired when compared to real images. This is because distribution of generated images can only be an approximation of the real distribution. In contrast, MISO extracts style features from real image $b$ to utilize information that is more complete.

\subsection{KL-Divergence Loss} \label{subsec:KL_loss}
At the test phase, we want to generate diverse outputs $G_{AB}(a, z)$ and $G_{BA}(b, z)$ where $z \sim \mathcal{N}(0, I)$. The KL-divergence loss encourages the encoded latent distribution $q(z_a|a)$ and $q(z_b|b)$ to be close to $\mathcal{N}(0, I)$, i.e.,
\begin{align}
\mathcal{L}_{KL}^{A} &= \mathbb{E}_{a \sim p(a)} [\mathcal{D}_{KL}(q(z_a|a) \thinspace || \thinspace \mathcal{N}(0, I))] \label{eq:KL_A_loss} \\ 
\mathcal{L}_{KL}^{B} &= \mathbb{E}_{b \thinspace \sim p(b)} \thinspace [\mathcal{D}_{KL}(q(z_b|b) \thinspace || \thinspace \mathcal{N}(0, I))] \label{eq:KL_B_loss}
\end{align}
Minimizing distance between only feature points will make it difficult for the model to handle unseen feature points but KL-divergence loss allows minimizing distances between two distributions. With KL-dviergence loss, our network can be more robust as an arbitrary $z$ sampled from $\mathcal{N}(0, I)$ will be trained to contain meaningful information about the target domain. The full KL-divergence loss is defined as
\begin{equation}
\mathcal{L}_{KL} = \mathcal{L}_{KL}^{A} + \mathcal{L}_{KL}^{B}
\end{equation}

\begin{figure*}
        \begin{subfigure}[b]{0.5\textwidth}
                \includegraphics[width=\linewidth]{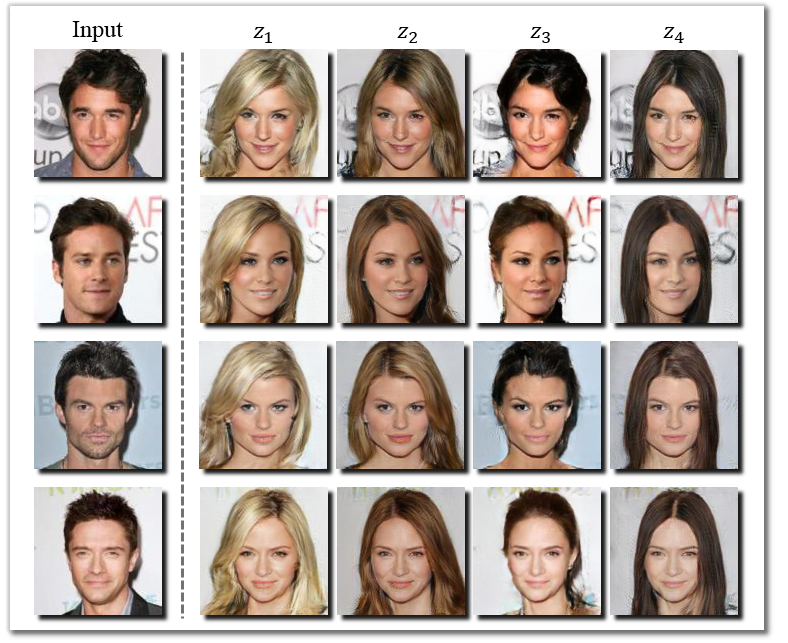}
                \caption{Male $\rightarrow$ Female}
                \label{subfig:Male2Female}
        \end{subfigure}%
        \begin{subfigure}[b]{0.5\textwidth}
                \includegraphics[width=\linewidth]{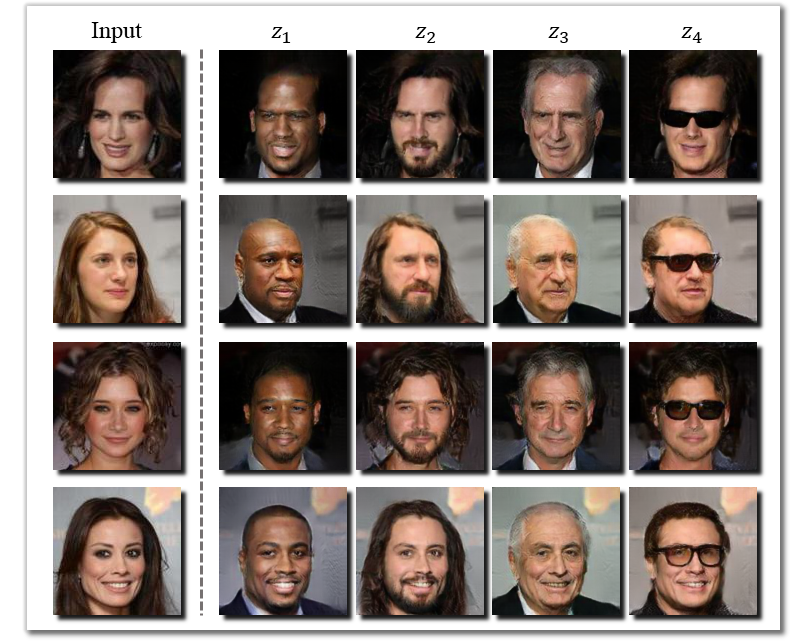} 
                \caption{Female $\rightarrow$ Male}
                \label{fig:Female2Male}
        \end{subfigure}%
        \caption{\textbf{Same $z$ for different source images.}. Each column shows images generated by the same random vector $z$. Different source images are translated to have the same features when given the same $z$, which shows that the model efficiently learns the domain-specific latent space. For instance in (a), $z_1$ corresponds to blond hair with a left side-part. Other random vectors also have their own unique domain-specific features such as brown hair, up-do hair style and light make-up.}\label{fig:same_sampling}
\end{figure*}

\subsection{Latent Reconstruction Loss} \label{subsec:lat_recon_loss}
We encourage the invertible mapping between $\mathcal{X}$ and $\mathcal{Z}$ with the latent reconstruction loss starting from $z\thinspace\sim\thinspace \mathcal{N}(0, I)$ which is a key part of the FIF phase in Fig. \ref{subfig:phase_2}. However, L1 loss between $z \sim \mathcal{N}(0, I)$ and $z_a\thinspace\sim\thinspace q(z_a|a)$ can be too strict and may bring instability to the training process. To avoid this, the latent reconstruction loss is defined as

\begin{equation} \label{eq:lat_recon_loss}
\mathcal{L}_{lat} = \mathbb{E} \thinspace _{b \sim p(b), \thinspace z \sim \mathcal{N}(0, I)}[|| \thinspace z - \mu_{A}^{out} \thinspace ||_1]
\end{equation}
where $\mu_{A}^{out}$ is one of the outputs from $E_A(G_{BA}(b, z))$.

\subsection{Full Objective Function} \label{subsec:full_loss}
Finally, we can formulate the full objective as 
\begin{align}
\mathcal{L}_{D} &= -\lambda_{adv} \mathcal{L}_{adv} \\
\mathcal{L}_{GE} &= \lambda_{adv} \mathcal{L}_{adv} + \lambda_{info} \mathcal{L}_{info} + \lambda_{cyc}\mathcal{L}_{cyc} \nonumber \\
& + \lambda_{KL}\mathcal{L}_{KL} + \lambda_{lat}\mathcal{L}_{lat}. 
\end{align}
Note that a same loss is also trained simultaneously with the two domains switched.

\section{Experiments}
To demonstrate the effectiveness of our model we conduct experiments on multiple datasets with various evaluation metrics and compare with other competitive baselines. The size of the images we used in our experiments is $128 \times 128$. 

\subsection{Datasets}
\textbf{Male $\leftrightarrow$ Female~~} CelebA~\cite{liu2015faceattributes} consisting of facial images with annotations. We separated all the images into two domains, male and female.

\textbf{Art $\leftrightarrow$ Photo~~} Monet $\leftrightarrow$ Photo dataset~\cite{CycleGAN2017} consisting of Monet's paintings and scenery photos.

\textbf{Summer $\leftrightarrow$ Winter~~}
Yosemite dataset~\cite{CycleGAN2017} consisting of summer scenes and winter scenes.

\textbf{Animal Translation~~}
Cat $\leftrightarrow$ Dog dataset~\cite{lee2018diverse} consisting of images with two kinds of dogs(husky and samoyed) and images with multiple kinds of cats.

\subsection{Baselines}
\textbf{CycleGAN+Noise (denoted as NycleGAN)~~}
To show the importance of two-phase training, we train a modified CycleGAN which injects noise vectors to its generator.

\textbf{MUNIT and DRIT~~}
These are state-of-the-art unpaired multimodal image-to-image translation models.

\textbf{SRSO~~}
To demonstrate that MILO is more effective than the SR loss, we train a variant of MISO that replaces MILO with a standard SR loss. Everything else is kept the same.

\textbf{w/o $\mathbf{L}_{\mathbf{adv}}^{\mathbf{rand}}$ and w/o $\mathbf{L}_{\mathbf{adv}}^{\mathbf{enc}}$~~}
These are variants of our model that have only one of our two losses, $L_{adv}^{rand}$ and $L_{adv}^{enc}$.

\begin{figure}
        \begin{subfigure}[b]{0.5\columnwidth}
                \includegraphics[width=\columnwidth]{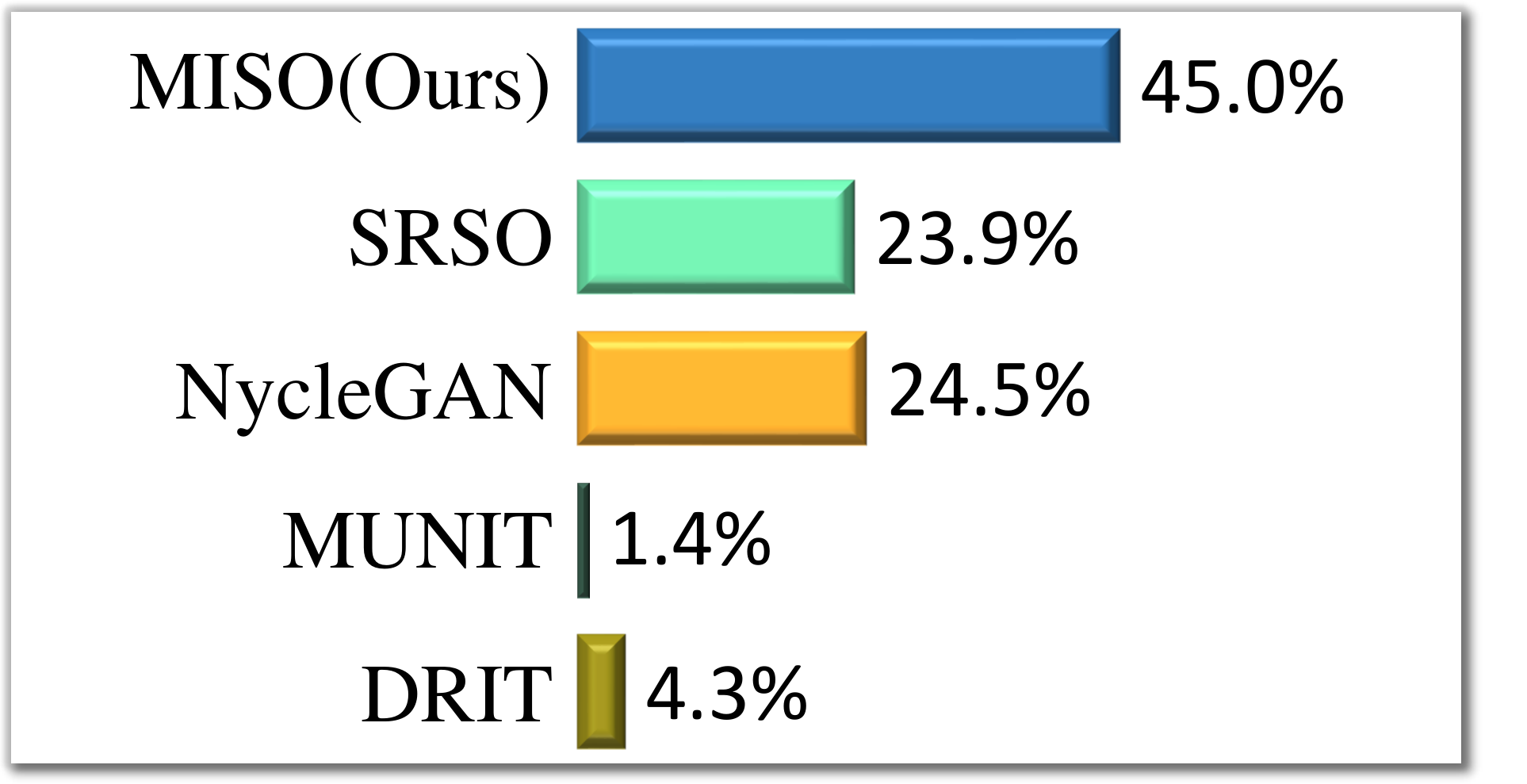}
                \caption{Male $\rightarrow$ Female}
                \label{subfig:userstudy_1}
        \end{subfigure}%
        \begin{subfigure}[b]{0.5\columnwidth}
                \includegraphics[width=\columnwidth]{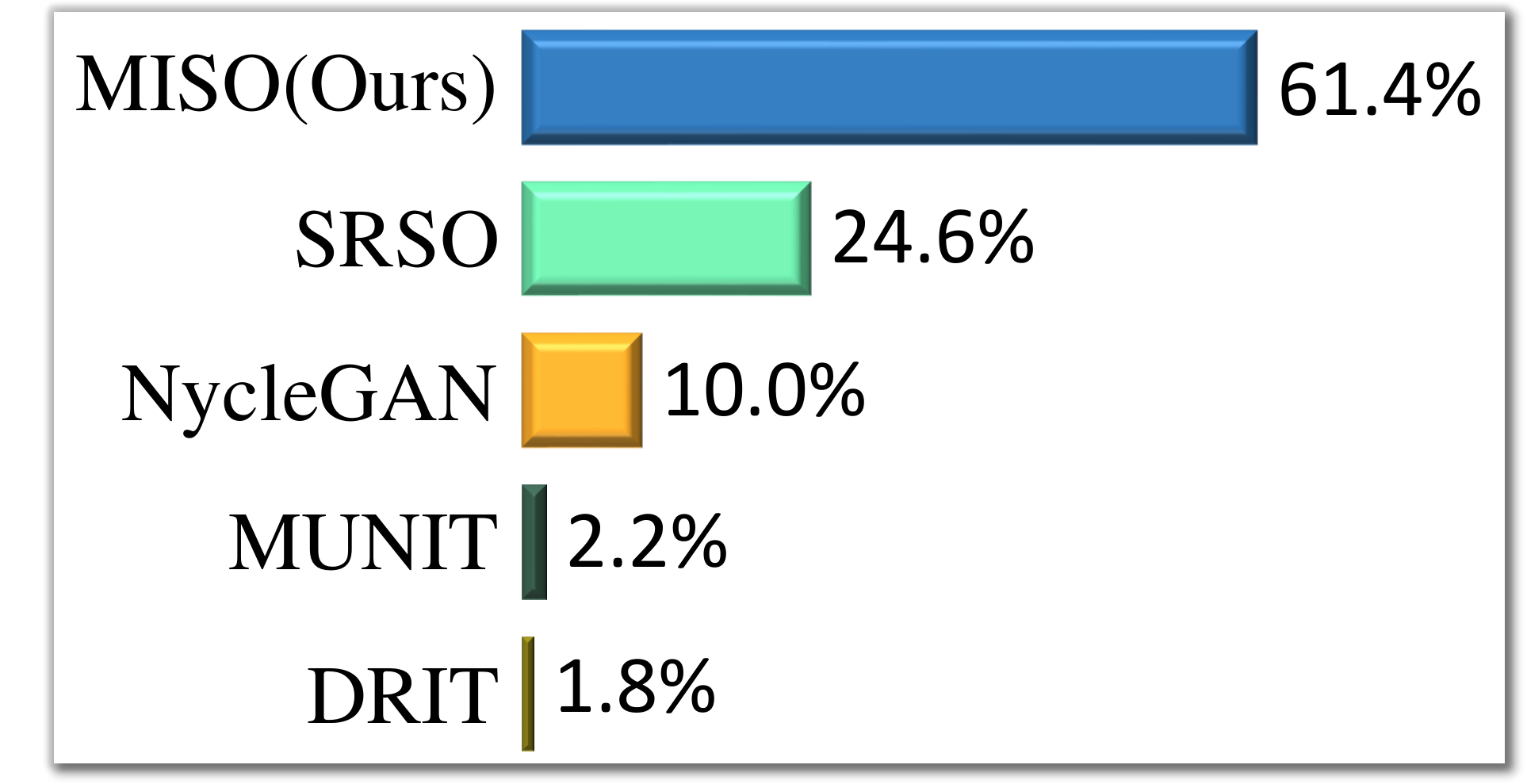}
                \caption{Female $\rightarrow$ Male}
                \label{subfig:userstudy_2}
        \end{subfigure}%
        \caption{\textbf{User preference on CelebA}. We conduct a user study for Male $\rightarrow$ Female and Female $\rightarrow$ Male on CelebA. 
        MISO shows superiority in both translations, especially Female$\rightarrow$Male in Fig.~\ref{subfig:userstudy_2}, which is consistent with the result in Table ~\ref{table:accuracy}.}\label{fig:userstudy}
\end{figure}
\vskip -0.07in
\begin{table}[t]
\vskip -0.1in
\caption{\textbf{Classification accuracy and likelihood of generated faces using CelebA.} A well-trained model should be able to get high scores on both accuracy and likelihood on a classifier trained on real data. As the standard of a well-trained model, we use a non-multimodal model, StarGAN~\cite{StarGAN2018}, which shows state-of-the-art performance on CelebA. In the first row, $\mathbb{F}$ and $\mathbb{M}$ denotes female and male.}
\label{table:accuracy}
\vskip -0.2in
\begin{center}
\begin{small}
\begin{sc}
\begin{tabular}{c|ccc}
\toprule
Model & Accuracy & $p(y=\mathbb{F}|x)$ & $p(y=\mathbb{M}|x)$ \\
\toprule
NycleGAN & 90.34 & 97.78 & 86.96 \\
DRIT    & 94.26 & 99.56 & 91.75 \\
MUNIT   & 98.40 & 99.58 & 97.89 \\
SRSO & 98.48 & 99.89 & 97.81 \\
\midrule
w/o $L_{adv}^{rand}$   & 92.61 & 99.59 & 89.45 \\
w/o $L_{adv}^{enc}$    & 96.87 & 98.34 & 96.19 \\
\midrule
\textbf{MISO(Ours)}    & 99.35 & 99.98 & 99.10 \\
\bottomrule
StarGAN & 98.81 & 99.93 & 98.28 \\
\bottomrule
\end{tabular}
\end{sc}
\end{small}
\end{center}
\vskip -0.2in
\end{table}

\begin{figure*}
        \begin{subfigure}[b]{0.5\textwidth}
                \includegraphics[width=\linewidth]{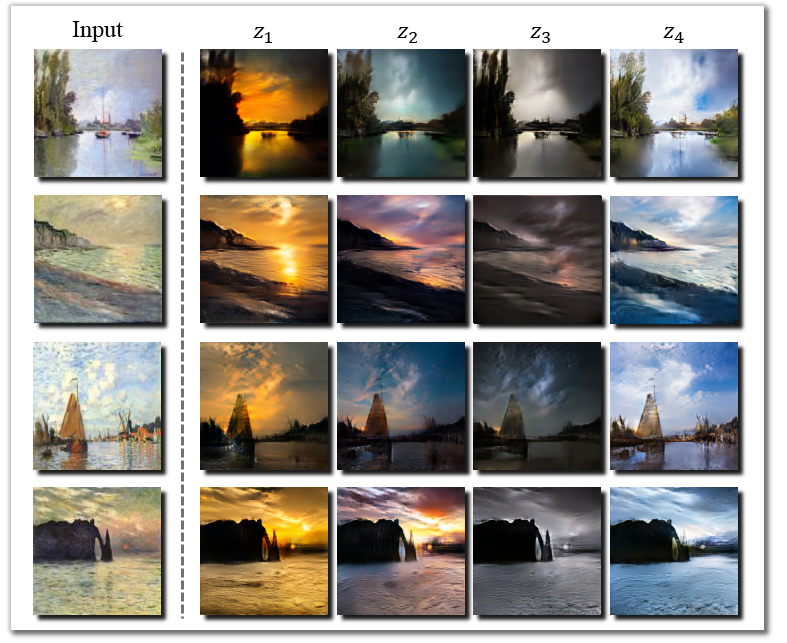}
                \caption{Monet $\rightarrow$ Photo}
                \label{subfig:Monet2Photo}
        \end{subfigure}%
        \begin{subfigure}[b]{0.5\textwidth}
                \includegraphics[width=\linewidth]{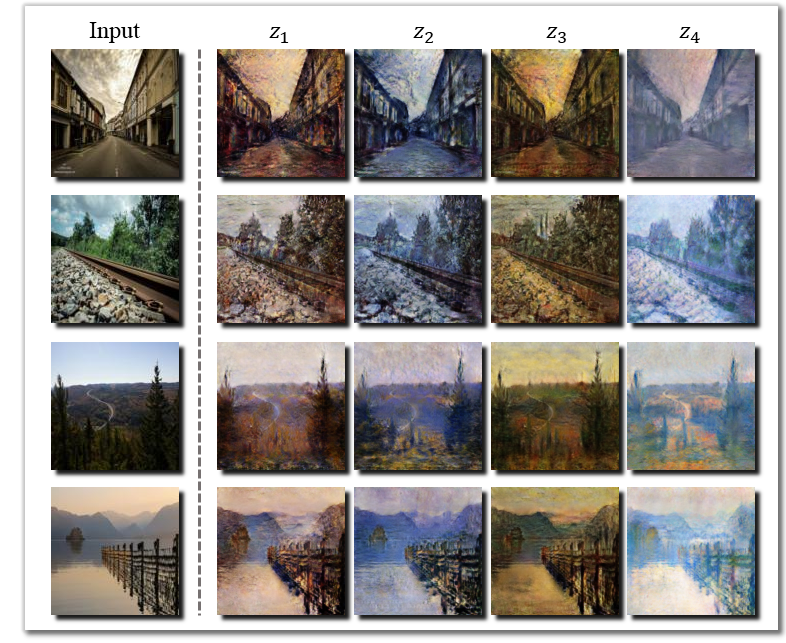} 
                \caption{Photo $\rightarrow$ Monet}
                \label{fig:Photo2Monet}
        \end{subfigure}%
        \caption{\textbf{Same $z$ for different source images.} All the random vectors have their own unique domain-specific features such as sunset, dark sunset, dark scene and bright scene in (a).}
        \vskip -0.1in
\end{figure*}

\begin{table}[t] 
\caption{\textbf{LPIPS distance in Monet $\leftrightarrow$ Photo translation}. Lower LPIPS between I(input)$\leftrightarrow$O(output) indicates that the output preserves content of the source image while higher LPIPS between O$\leftrightarrow$O means that outputs are more diverse.}
\label{table:LPIPS}
\vskip 0.15in
\begin{center}
\begin{small}
\begin{sc}
\begin{tabular}{c|cc|cc}
\toprule
  & \multicolumn{2}{c|}{Monet $\rightarrow$ Photo} & \multicolumn{2}{c}{Photo $\rightarrow$ Monet}\\
\midrule
Model &  I $\leftrightarrow$ O & O $\leftrightarrow$ O & I $\leftrightarrow$ O & O $\leftrightarrow$ O \\
\toprule
NycleGAN    & 0.4052 & 0.0044 & 0.5946 & 0.0025 \\
DRIT    & 0.5829 & 0.2838  & 0.6572 & 0.3807 \\
MUNIT   & 0.6571 & 0.5947  & 0.6154 & 0.4786 \\
SRSO    & 0.5168 & 0.4798  & 0.6103 & 0.5150 \\
\midrule
w/o $\mathcal{L}_{adv}^{rand}$   & 0.7553 & 0.3660 & 0.7523 & 0.3957 \\
w/o $\mathcal{L}_{adv}^{enc}$    & 0.5515 & 0.5102 & 0.5616 & 0.4764 \\
\midrule
\textbf{MISO(Ours)}    & 0.4549 & 0.3889  & 0.4811 & 0.3887 \\
\bottomrule
real    & N/A & 0.3981 & N/A & 0.3984 \\
\bottomrule
\end{tabular}
\end{sc}
\end{small}
\end{center}
\vskip -0.1in
\end{table}

\vskip -0.1in

\subsection{Evaluation Metric}
\textbf{User Preference~~}
To compare realism and quality of translation outputs of various models, we perform a user study with 30 participants. Each participant answered 40 questions in total (20 for male$\rightarrow$female translation and 20 for female$\rightarrow$male). We give a random source image and its corresponding generated outputs of our model and baselines. We then ask which generated output has the highest quality while maintaining content of the source image. 

\begin{figure} 
\begin{center}
\includegraphics[width=1\columnwidth]{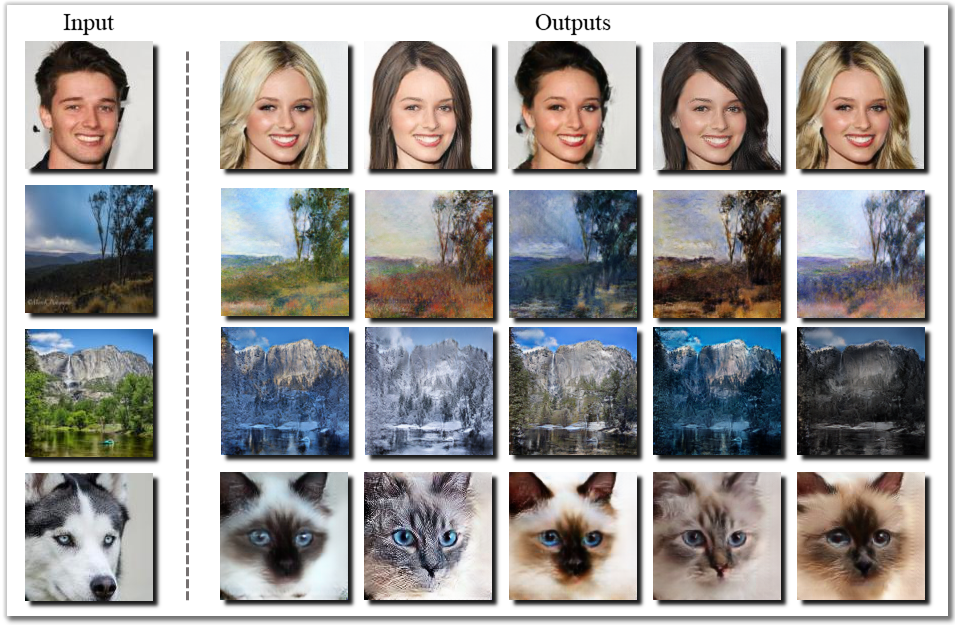}
\end{center}
\vskip -0.2in
\caption{\textbf{Samples generated with random $z$}. From the top: male$\rightarrow$female, photo$\rightarrow$Monet, summer$\rightarrow$winter and dog$\rightarrow$cat. MISO is capable of generating diverse and high-quality outputs.}
\label{fig:random_sampling}
\vskip -0.1in
\end{figure} 

\textbf{Classification Accuracy and Likelihood~~}
We measure realism of generated outputs by the classification accuracy and the likelihood of images translated on CelebA~\cite{liu2015faceattributes} using a classifier trained on real data. A successful transfer would result in a male image transferred to a female image being classified as female, and vice versa. 10 different images are generated for each of 300 input images.

\begin{figure*}
    \begin{center}
    \includegraphics[width=0.9\linewidth]{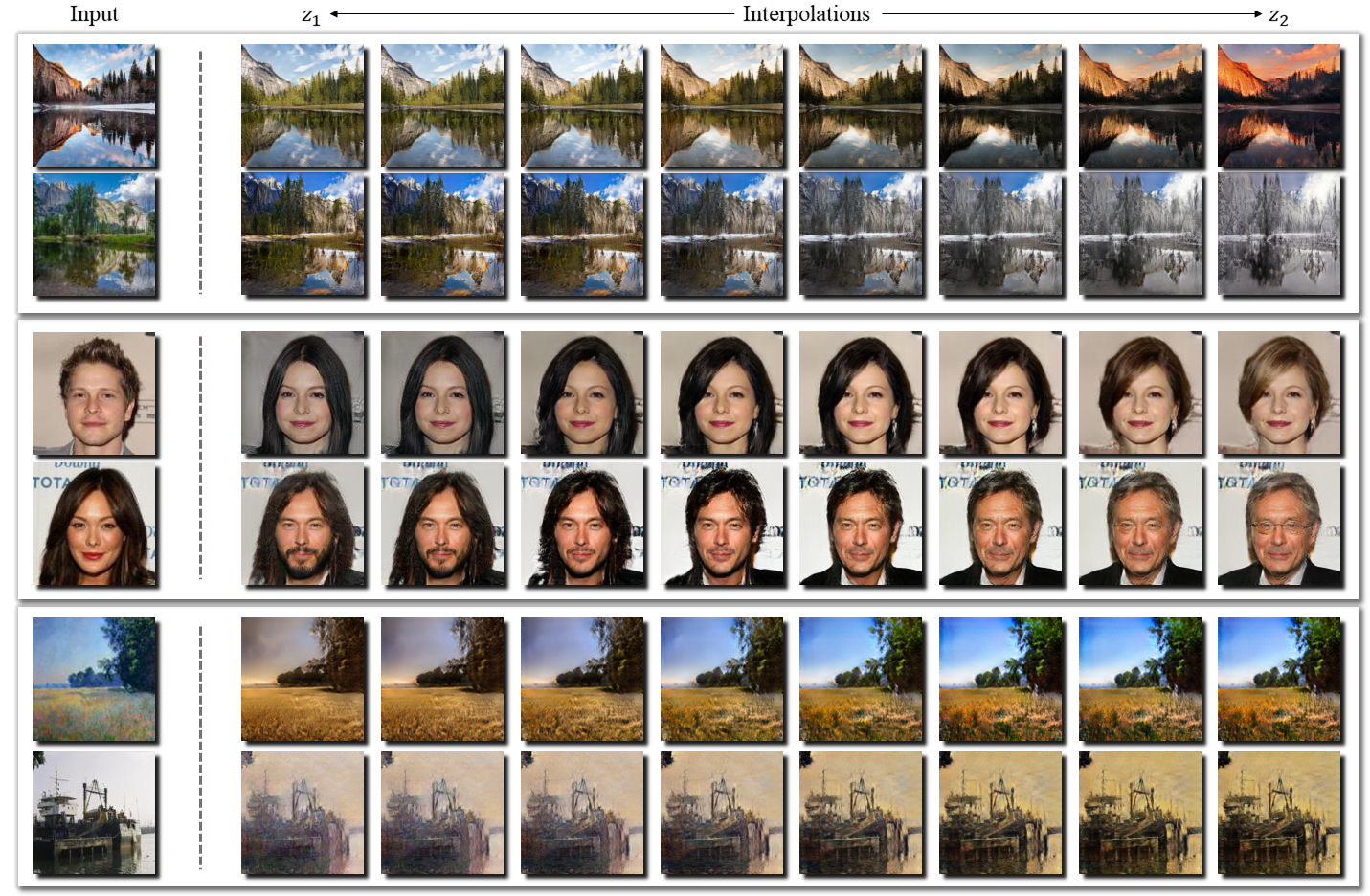}
    \end{center}
    \vskip -0.2in
    \caption{\textbf{Translation using vectors interpolated between two random vectors}. From the top: winter$\rightarrow$summer, summer$\rightarrow$winter, male$\rightarrow$female, female$\rightarrow$male, monet$\rightarrow$photo and photo$\rightarrow$monet. We observe that our model successfully translates unique domain-specific features encoded in each random vector into continuous and diverse outputs that slide smoothly into different features.}
    \label{fig:interpolation_sampling}
\end{figure*}

\textbf{LPIPS Distance~~}
LPIPS~\cite{zhang2018unreasonable} measures perceptual distance between images to mimic human perceptual similarities. We measure two distances: 1) Similarity of translated images to an input image are measured by average LPIPS between the input image and its corresponding translated images($I \leftrightarrow O$ distance) 2) Diversity of translated images are measured by calculating average LPIPS distance between images generated from the same source image($O \leftrightarrow O$ distance). Average distance of real-world data is set as an upper bound of $O \leftrightarrow O$ distance to discern whether the model makes realistic diversities. 10 different images are generated for each of 800 input images.

\vskip -0.1in

\section{Results}
This section reports results on quantitative experiments and qualitative results to show the effectiveness of our model in terms of realism and diversity.

\subsection{Quantitative Results}
We perform quantitative analysis of realism, content preservation, and diversity of MISO in comparison to baselines. 

\textbf{Realism} In shown in Table~\ref{table:accuracy}, our model exhibits the highest gender classification accuracy for both female and male images transferred from a source image of the opposite gender.
Interestingly, baseline models struggle in female $\rightarrow$ male conversion as seen from the lower likelihood of the right column in Table~\ref{table:accuracy}. In contrast, our model shows less than one percent error rate, which implies that it is well trained on conversion to both domains and produces realistic images that contain domain-specific features.

\textbf{Content Preservation and Diversity}
In Table~\ref{table:LPIPS}, our model has the lowest I$\leftrightarrow$O distance, implying that it works best in preserving content of the source image, excluding NycleGAN. NycleGAN may preserve content but it cannot produce diverse outputs. Regarding diversity of outputs, our model has the highest O$\leftrightarrow$O diversity score under the upper bound. MUNIT and SRSO show the high diversity score in each translation of Monet $\rightarrow$ Photo and Photo $\rightarrow$ Monet but also has a high I$\leftrightarrow$O score which implies that its outputs are diverse but unrealistic. Thus, our full model best achieves diversity without a significant trade-off of realism. 

\subsection{Qualitative Results}

\textbf{Comparison to Baselines on Monet Paintings~~}
In Fig.~\ref{subfig:comparison_sub_1}, we can see that the baseline models fail at maintaining contents of the source image. DRIT generates white spots on the place where the tree in the source image should be, and MUNIT generates distorted unrealistic images. 
A variation of our model, SRSO, which replaces MILO with SR loss, maintains the source image better than both DRIT and MUNIT. It manages to generate a shape similar to the tree in the source image. While other baseline models strictly separate content and style in the disentangling process, our model considers the content as a base and adds style as a condition to perform translation.
This could be the reason our model succeeds in maintaining the content of the source image. Our full model with MILO succeeds in maintaining the details of trees and ground in the source image while generating the most realistic and diverse images (e.g., sunset in the first row, scenes that look like summer and fall in the second and the third rows, respectively). It is consistent with the result of the quantitative experiment on content preservation and diversity in Table~\ref{table:LPIPS}.

\textbf{Comparison to Baselines on Face Images~~} 
Compared to images such as paintings and animals, generated faces may need higher image quality because people are generally sensitive to visual artifacts in human faces~\cite{chang2018pairedcyclegan}. In Fig.~\ref{subfig:comparison_sub_2}, DRIT is able to make differences only in terms of overall color tones among the generated images. Also, most of the generated hair regions are blurry, which is undesirable as hair is one of the most important features that differentiate the female domain. This is mainly because of the limitation of SR loss. Although MUNIT generates more diverse images than DRIT, the generated images are not of high quality. For example, images in the first row and third row fails to maintain the background, and the details around shoulders are somewhat poor. These results show the importance of treating latent representations as a random variable and making its distribution close to $\mathcal{N}(0, I)$. In MUNIT, the latent variables are considered deterministic, so KL-divergence loss is not applicable. Instead, MUNIT relies on the latent reconstruction loss which can cause problems for generating images with unseen $z \sim \mathcal{N}(0, I)$ because it minimizes the distance between feature points not distributions. Although SRSO generates more diverse images than DRIT and more clear images than MUNIT, the generated images are still blurry around the hair(first and third rows), produces visual artifacts (second and third rows) and even sometimes collapse (the second row). Our full model MISO produces realistic but diverse outputs by diversifying detailed features such as hair color and style as well as make-up style. The generated images of high quality in Fig.~\ref{subfig:comparison_sub_2} support the superiority of our model.

\textbf{Learning the Latent Space.} 
In Fig~\ref{fig:interpolation_sampling}, we can see that our model learns the data distribution and the latent space effectively instead of just simply memorizing the training data. 


\section{Conclusion}
In this paper, we presented a novel framework of unpaired multimodal image-to-image translation that achieves state-of-the-art performance on various datasets utilizing conditional encoder as well as a new mutual information loss function. Our model generates high-quality and diverse images with meaningful characteristics while preserving the content of the source image. The style-conditioned content and information-theoretic loss function motivated by the interpretation of latent variables as a random variable results in superiority of our model.

\bibliography{egbib}
\bibliographystyle{icml2019}

\end{document}